\documentclass[letterpaper]{article} 
\usepackage{aaai25}  
\usepackage{times}  
\usepackage{helvet}  
\usepackage{courier}  
\usepackage[hyphens]{url}  
\usepackage{graphicx} 
\usepackage{natbib}  
\usepackage{caption} 
\usepackage{algorithm}
\usepackage{algorithmic}

\usepackage{newfloat}
\usepackage{listings}
\DeclareCaptionStyle{ruled}{labelfont=normalfont,labelsep=colon,strut=off} 
\floatstyle{ruled}
\newfloat{listing}{tb}{lst}{}
\floatname{listing}{Listing}
\usepackage{graphicx}
\usepackage{amsmath}
\usepackage{multirow}
\usepackage{diagbox}
\usepackage{subcaption}
\usepackage{amsfonts}
\usepackage{booktabs}

\title{A New Adversarial Perspective for LiDAR-based 3D Object Detection}
\author{
    Shijun Zheng\textsuperscript{\rm 1,\rm 2},
    Weiquan Liu\textsuperscript{\rm 3}\thanks{Corresponding Author.},
    Yu Guo\textsuperscript{\rm 1,\rm 2},
    Yu Zang\textsuperscript{\rm 1,\rm 2}\footnotemark[1],
    Siqi Shen\textsuperscript{\rm 1,\rm 2},
    Cheng Wang\textsuperscript{\rm 1,\rm 2}
}
\affiliations{
    \textsuperscript{\rm 1}Fujian Key Laboratory of Sensing and Computing for Smart Cities, School of Informatics, Xiamen University, China\\
    \textsuperscript{\rm 2}Key Laboratory of Multimedia Trusted Perception and Efficient Computing, Ministry of Education of China,\\
    School of Informatics, Xiamen University, China\\
    \textsuperscript{\rm 3}College of Computer Engineering, Jimei University, China\\
    \{zhengshijun, gouyu\}@stu.xmu.edu.cn, \{siqishen, cwang\}@xmu.edu.cn, wqliu@jmu.edu.cn, zangyu7@126.com
}

\usepackage{bibentry}

\begin{document}

\maketitle

\begin{abstract}
Autonomous vehicles (AVs) rely on LiDAR sensors for environmental perception and decision-making in driving scenarios. However, ensuring the safety and reliability of AVs in complex environments remains a pressing challenge. To address this issue, we introduce a real-world dataset (ROLiD) comprising LiDAR-scanned point clouds of two random objects: water mist and smoke. In this paper, we introduce a novel adversarial perspective by proposing an attack framework that utilizes water mist and smoke to simulate environmental interference. Specifically, we propose a point cloud sequence generation method using a motion and content decomposition generative adversarial network named PCS-GAN to simulate the distribution of random objects. Furthermore, leveraging the simulated LiDAR scanning characteristics implemented with Range Image, we examine the effects of introducing random object perturbations at various positions on the target vehicle. Extensive experiments demonstrate that adversarial perturbations based on random objects effectively deceive vehicle detection and reduce the recognition rate of 3D object detection models.
\end{abstract}


\section{Introduction}
LiDAR is an important support for modern autonomous driving systems in achieving scene perception due to its ability to capture accurate depth information~\cite{xia2021vpc}. Learning-based LiDAR point cloud methods~\cite{xia2023casspr, xia2024text2loc} have recently become popular with the development of deep learning. However, previous studies have shown that deep neural network models are vulnerable to adversarial attacks, causing the model to produce erroneous outputs \cite{szegedy2014intriguing,huang2022shape}. If an attacker conducts malicious attacks on deep network models, it will pose a great threat to the security of the autonomous driving system.

Currently, 3D adversarial attacks are primarily focused on the digital domain, often overlooking the potential for adversarial attacks in real-world scenarios. In the digital domain, adversarial attack methods include perturbing points, adding independent points or adversarial clustering ~\cite{xiang2019generating}, and deleting points \cite{zheng2019pointcloud}. In addition, using gradients to implement adversarial attacks is also a major method ~\cite{ma2020efficient}, but the generated adversarial examples are prone to outliers. Optimization-based methods~\cite{zheng2023adaptive,wen2020geometry} can generate adversarial examples with better geometric properties, but this method performs slowly. In the frequency domain, the imperceptibility and transferability of adversarial examples can be improved based on high-frequency and low-frequency information. However, the adversarial examples generated by these methods are not operable in the real world.

\begin{figure}
	\centering 
	\includegraphics[width=0.98 \linewidth]{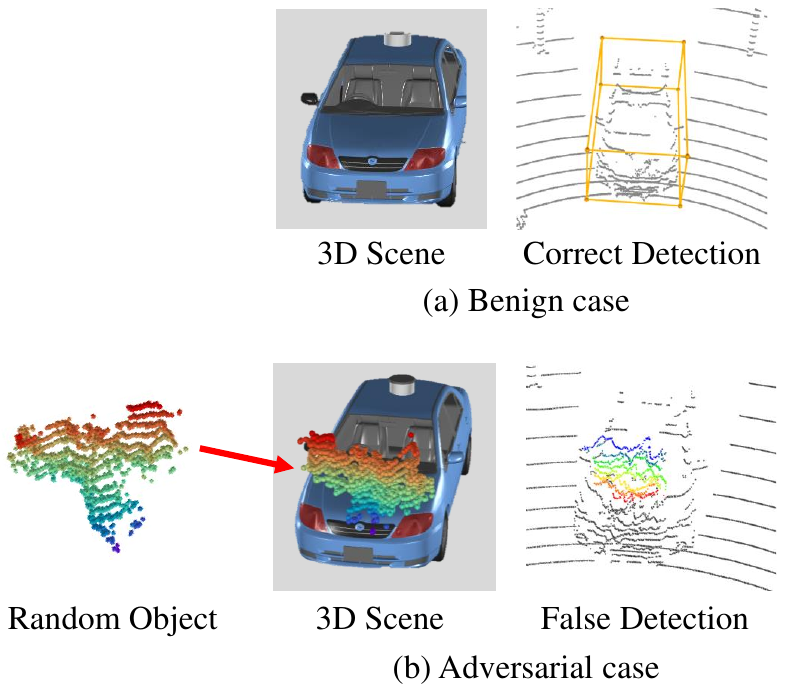}
	\caption{We use random objects to camouflage the target vehicle and fool the LiDAR detector. In benign cases, the 3D object detector correctly detects the vehicle. In adversarial cases, the detector fails to identify the vehicle when random objects are superimposed on it.}
	\label{fig:intro}
\end{figure}

The methods for addressing adversarial attacks in real-world scenarios can be divided into two categories. The first is a sensor-level attack. Attackers use malicious laser pulses to simulate LiDAR return pulses and inject attack signals into LiDAR signals ~\cite{cao2019adversarial}. The second type is a model-level LiDAR attack, which is also the main adversarial attack method. For example, adding adversarial objects to the roof of a vehicle makes it invisible to LiDAR detectors, thereby deceiving the detection system~\cite{tu2020physically}. Adversarial objects are generated optimally using adversarial attack methods in the digital domain, and then 3D printing is used to produce this adversarial example in the real world. Such adversarial examples placed on real roads can mislead autonomous vehicles equipped with LiDAR~\cite{cao2021invisible}. These adversarial attack methods seek to generate optimal adversarial objects, which generally have a fixed appearance and are not prone to deformation, such as geometric objects~\cite{tu2020physically}, traffic cones~\cite{cao2021invisible}, etc.

In this paper, different from existing 3D adversarial attacks on real scenes, we focus on the adversarial nature of random objects to LiDAR. We explore the adversarial capabilities of two random objects: water mist and smoke. To this end, we introduce an adversarial attack framework based on random object interference from a new perspective. First, since random objects have complex physical meanings, they are difficult to describe quantitatively and qualitatively. Therefore, we designs a motion and content separation generative adversarial network method for point cloud sequence generation. This method is used to generate point cloud sequences of random objects, which simulate the distribution of random objects during LiDAR scanning. Second, to verify the attack capability of random objects, we use different methods to superimpose simulated random objects on different positions of vehicle targets in public autonomous driving datasets to generate adversarial data. These generated adversarial data are then used to attack state-of-the-art 3D object detection methods. 

Overall, the contributions of this paper can be summarized as follows:
\begin{itemize}
	\item We are the first to delve into the adversarial nature of random objects for LiDAR perception in real-world scenarios. By employing two random objects, namely water mist and smoke, we demonstrate adversarial attacks on 3D object perception in real scenes.
	\item  We are the first to propose a generative adversarial network framework (PCS-GAN) for point cloud sequence generation. This framework is used to generate random objects that simulate the data characteristics of LiDAR scanning.
	\item We construct a LiDAR point cloud dataset (ROLiD) of random objects, including water mist and smoke data. The dataset will be released for public research.
	\item Our method effectively attacks state-of-the-art 3D detectors on KITTI and nuScenes, with attack success rates greater than 80\% for most models.
\end{itemize}

\section{Related Work}
\textbf{Adversarial attacks in the digital domain.} The 3D point cloud adversarial attack method was originally extended from the image adversarial attack method. Attackers seek to generate adversarial examples with high attack success rates and better imperceptibility. Xiang et al. ~\cite{xiang2019generating} conducted pioneering research on adversarial attacks targeting 3D point clouds, extending the $C\&W$ attack to this domain. Wen et al. ~\cite{wen2020geometry} argued that regularization terms used to constrain perturbation size fail to ensure imperceptibility. To address this limitation, they proposed a local curvature consistency measure for evaluating point cloud similarity from a geometric perspective and developed a geometry-aware attack method. In addition, generating high-quality adversarial samples with minimal cost remains a crucial research focus~\cite{kim2021minimal,zheng2023adaptive}. For example, Zheng et al.~\cite{zheng2023adaptive} introduced a local region adversarial attack method on 3D point clouds. In recent years, to improve the transferability and imperceptibility of adversarial examples, researchers have focused increasingly on attacks in the graph spectral domain~\cite{hu2022exploring,tao20233dhacker}. Hu et al.~\cite{hu2022exploring} proposed a new point cloud attack paradigm, namely graph spectral domain attack (GSDA), which generates adversarial samples by perturbing the transformation coefficients corresponding to different geometric structures in the graph spectral domain.

\textbf{Adversarial attacks in real scenarios.} Implementing adversarial attacks in real-world settings is notably more complex and presents significant security challenges for autonomous driving systems. Tu et al.~\cite{tu2020physically} proposed an adversarial attack method to generate adversarial objects of different geometric shapes, and placing them on top of the car can fool the LiDAR detector, causing the car to be invisible. Cao et al.~\cite{cao2021invisible} explored the security vulnerabilities of multi-sensor autonomous driving systems by formulating adversarial attacks as an optimization problem. To this end, the authors introduced MSF-ADV~\cite{cao2021invisible} to generate physically realizable, adversarial 3D-printed objects that can mislead autonomous driving systems. The attack algorithm was evaluated using an industrial-grade autonomous driving system in the real world and achieved a success rate of more than 90\%. Considering the geometric characteristics of 3D objects and the invariance of physical transformation, Miao et al.~\cite{dong2022isometric} introduced Gaussian curvature into the regularization term to generate natural and robust 3D adversarial samples in the physical world. Zhu et al.~\cite{zhu2021can} proposed an adversarial attack method to find specific attack locations in the real world. Placing any object with a reflective surface, such as a commercial drone, at these attack locations can mislead LiDAR sensing systems. This method is also the first to study the impact of adversarial position on LiDAR perception models.

\begin{figure*}
	\centering 
	\includegraphics[width=1.0 \linewidth]{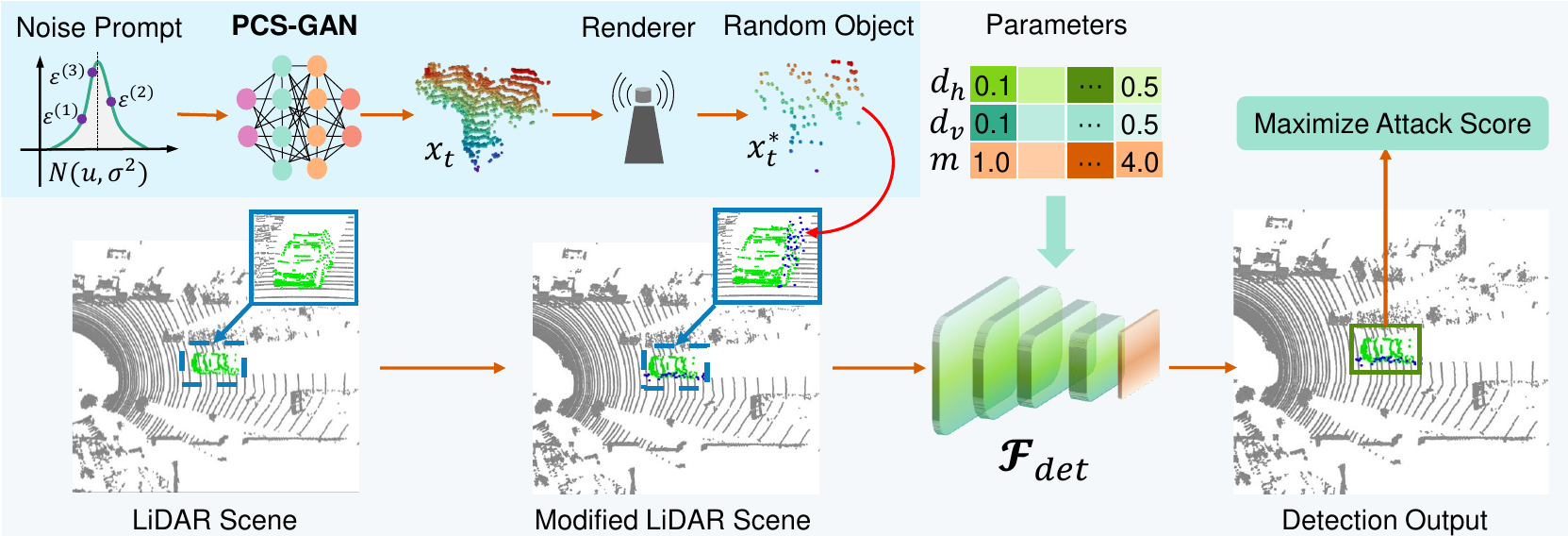}
	\caption{Our adversarial attack framework based on random object perturbations. We propose PCS-GAN to generate random object sequences ($\widetilde{P}_{T}=[\widetilde{x}^{(1)},\widetilde{x}^{(2)},\cdot \cdot \cdot,\widetilde{x}^{(K)}]$) and use a range image representation approximate renderer to simulate LiDAR scanning. We attach random objects to the target vehicle by setting different fusion modes ($m$) and fusion densities ($d_h,d_v$) to maximize the attack score of the target vehicle on LiDAR detector ($\mathcal{F}_{det}$).}
	\label{fig:teaser}
\end{figure*}

In summary, compared with existing adversarial attack methods in the real scenarios, our work is oriented towards adversarial attacks on random objects such as water mist and smoke. We generate random objects through algorithms and simulate their distribution. We do not use optimization methods to find optimal adversarial objects. Such adversarial objects have a fixed geometric appearance and are not easily deformed. At the same time, we explore the attack performance when adding random object perturbations at different locations of the target object. In the process of adding perturbation, we consider the occlusion problem of the target object and simulated the physical characteristics of real LiDAR scanning.

\section{Method}

\subsection{Problem Formulation}
In this paper, we aim to use random objects as camouflage to conduct adversarial attacks on 3D point clouds in real scenarios. We formulate the problem of attacks based on random object perturbations into two parts, illustrated in Figure \ref{fig:teaser}. The first part is the generation of random objects. We use the real random object $y^{train}$ to train the deep model $\mathcal{F}$ to obtain the optimal parameters $\theta^*$ when the noise $r^{train}$ is triggered, as follows:
\begin{equation}
	\begin{aligned}
		& \mathcal{F}_{\theta^*}=\underset{\theta}{\arg \min }\left(\mathcal{L}^{\text {full }}\left(\mathcal{F}_\theta\left(\boldsymbol{r}^{\text {train }}\right), \boldsymbol{y}^{\text {train }}\right)\right).
	\end{aligned}
\end{equation}

The second part is the generation and optimization of adversarial examples. We use the optimal random objects generated by noise $r^{val}$ under the optimal parameters of network $\mathcal{F}$ and fuse them with the target vehicle into adversarial point clouds. Subsequently, the attack score of the target vehicle $s^{val}$ is maximized on the LiDAR detector $\mathcal{F}_{det}$ by optimizing different fusion modes $m$ and fusion densities $d$ as follows:
\begin{equation}
	\begin{aligned}
		\mathcal{F}_{\theta^*}^{\eta^*}=\underset{\eta}{\max }\left(\mathcal{F}_{det}\left(\mathcal{F}_{\theta^*}^\eta\left(\boldsymbol{r}^{\text {val }}\right), \boldsymbol{s}^{\text {val }}\right)\right),
	\end{aligned}
\end{equation}
where parameter $\eta=[m,d]$.

\subsection{Dataset of random objects}
In this paper, we present a LiDAR point cloud dataset (ROLiD) of random objects for the study of data simulation and adversarial attacks in the real world.

\begin{figure}
	\centering 
	\includegraphics[width=1.0 \linewidth]{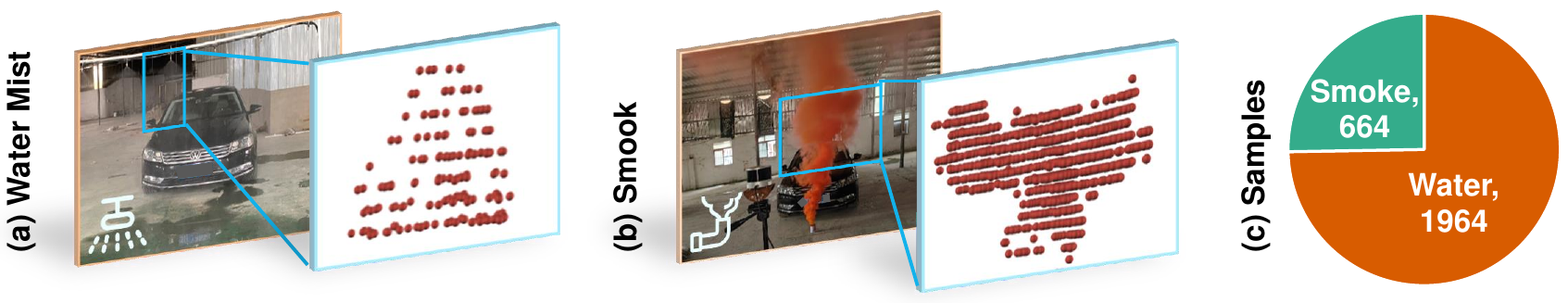}
	\caption{Random objects LiDAR Dataset (ROLiD). (a) and (b) are the data collection scheme and local point cloud visualization respectively. (c) represents the data used for the PCS-GAN network.}
	\label{fig:data}
\end{figure}

\textbf{Data Acquisition.} We use 32-line LiDAR to collect data, including water mist data and smoke data as shown in Figure \ref{fig:data}. The collection environment is an open factory building, and the supplementary material explains the collection equipment and detailed data acquisition methods.

\textbf{Water mist data.} We spray water mist in the four directions of the vehicle's head, tail, left, and right, and use LiDAR to scan the point cloud directly against the vehicle, as shown in Figure \ref{fig:data}(a). When collecting water mist point clouds in each direction of the vehicle, the LiDAR is fixed 10 meters away from the vehicle, and four intensities of water pressure are used, namely 0.45Mpa, 0.50Mpa, 0.55Mpa, and 0.60Mpa. Under each water pressure, we set the distance between the water mist position and the car to 2 meters, 4 meters, 6 meters, and 8 meters, respectively. The collection time of water mist point clouds at each distance under each pressure is about 3 minutes. The total number of water mist data exceeds 128k frames.

\textbf{Smoke data.} We released smoke near the vehicle in four directions: the front, rear, left, and right sides. LiDAR was then used to scan the point cloud directly from these positions, as shown in Figure \ref{fig:data}(b). The smoke was emitted for approximately 2-3 minutes.

\textbf{Data Characteristic.}
We quantified the impact of water mist data on vehicles to demonstrate the adversarial nature of random objects. Under the same conditions for collecting water mist data, we use LiDAR to collect target vehicle point clouds without spraying water mist. To quantify the impact of the water mist on the vehicle, We counted the number of vehicle point cloud before and after it obscuration by water mist and calculated the occlusion ratio. The quantitative results for different distances under different pressures can be found in supplementary material.

\subsection{Sequence generation of random objects}
Random objects such as water mist and smoke have different distribution patterns depending on the surrounding environment. To better simulate the distribution of random objects during LiDAR scanning, we propose a novel Point Cloud Sequence generation method using a Generative Adversarial Network with a motion and content decomposition strategy (PCS-GAN).

Separating content and motion is an effective method for video generation~\cite{tulyakov2018mocogan}. Similarly, for the problem of point cloud sequence generation, the motion information between sequences can be separated from the content of the point cloud itself. Therefore, we can decompose the point cloud latent space $Z$ into content space $Z_C$ and motion space $Z_M$. For random objects, the motion space $Z_M$ is used to model the changing characteristics between point cloud sequences, and the content space $Z_C$ is used to model the invariant characteristics of the point cloud.

\begin{figure}
	\centering
	\includegraphics[width=1.0 \linewidth]{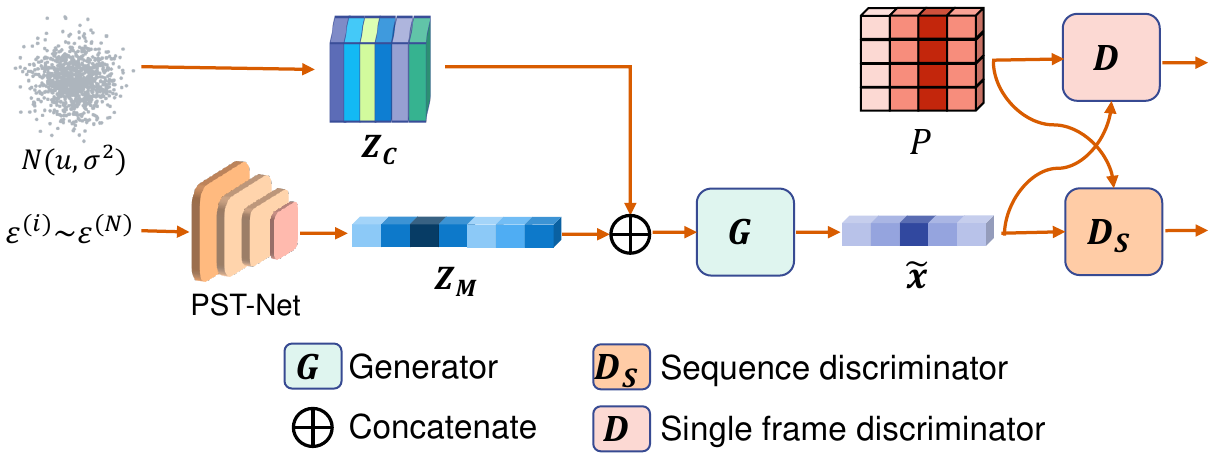}
	\caption{Generative adversarial network framework PCS-GAN for point cloud sequence generation. $P$ is real point cloud, $\widetilde{x}$ is fake point cloud , $Z_C$ is content features, $Z_M$ motion features.}
\label{fig:GAN}
\end{figure}

The proposed PCS-GAN consists of four components, namely PST-Net, generator $G$, single-frame discriminator $D$, and sequence discriminator $D_S$, as shown in Figure \ref{fig:GAN}.

We implement a feature extraction network, PST-Net, based on point spatio-temporal convolution~\cite{fan2022pstnet} to extract temporal and spatial features of point cloud sequences. The input of the network is a point cloud sequence $\varepsilon = (\varepsilon^{(1)},\varepsilon^{(2)}, \cdot \cdot \cdot,\varepsilon^{(N)})$ obeying Gaussian distribution, and the output is the information representation $Z_M$ of the point cloud sequence.

The goal of the generator $G$ is to generate point clouds with similar statistical distributions to the real point clouds. Considering the dynamic changing characteristics of random objects, in order to better learn the complex shapes of objects, we use Warping-based Generator~\cite{tang2022warpinggan}. The motion information $Z_M$ and content information $Z_C$ of the point cloud sequence are concatenated as the input of the generator, and the output is a sequence of K-frame point clouds.

The discriminators $D$ and $D_S$ are used to determine whether the point cloud generated by the generator $G$ is a real point cloud. Among them, the discriminator $D$ is used to determine whether the single-frame point cloud of the point cloud sequence generated by the generator $G$ is a real point cloud. We adopt PointNet~\cite{qi2017pointnet} as the single-frame point cloud discriminator $D$. For generating point clouds, the input to the discriminator is a randomly sampled frame from the point cloud sequence generated by generator $G$. $D_S$ is a temporal discriminator~\cite{li2021tpu} of a point cloud sequence, which is used to determine whether the point cloud sequence generated by the generator $G$ is a real point cloud. We sample three consecutive frame point clouds from the point cloud sequence generated by generator $G$ as input to predict the temporal consistency confidence of the point cloud sequence. Similarly, for the real point cloud sequence, three consecutive frames are used as the input of $D_S$.

To generate a point cloud sequence, a random vector $\varepsilon = (\varepsilon^{(1)},\varepsilon^{(2)}, \cdot \cdot \cdot,\varepsilon^{(N)})$ of length $N$ is sampled from the Gaussian distribution. After the motion information is extracted through the PST-Net network, the sequence length will be reduced to K, recorded as $Z_M = (Z_{M}^{1},Z_{M}^{2}, \cdot \cdot \cdot,Z_{M}^{K})$. Therefore, a random vector of length $K$ is sampled from a Gaussian distribution and concated with $Z_M$ as the input to the generator $G$. The sequence generated by the generator $G$ is marked as $\widetilde{P}_{T}=[\widetilde{x}^{(1)},\widetilde{x}^{(2)},\cdot \cdot \cdot,\widetilde{x}^{(K)}]$, from which we randomly sample a frame of point clouds as $\widetilde{P}=[\widetilde{x}^{(i)}]$, and the point cloud sequence of three consecutive frames is marked as $\widetilde{P}_{T}=[\widetilde{x}^{(i-1)},\widetilde{x}^{(i)},\widetilde{x}^{(i+1)}]$. Similarly, a frame randomly sampled from the real point cloud is recorded as ${P}=[x^{(i)}]$, and a point cloud sequence of three consecutive frames is recorded as $P_{T}=[x^{(i-1)},x^{(i)},x^{(i+1)}]$.

We mark the PST-Net network as $P_M$. Therefore, according to GAN~\cite{creswell2018generative} and the video generation framework MoCoGAN~\cite{tulyakov2018mocogan}, the learning problem of PCS-GAN is expressed as follows:
\begin{equation}
\max _{G, P_{M}} \min _{D,D_{S}} \mathcal{F}_{\mathrm{P}}\left(D, D_{\mathrm{S}}, G,  P_{M} \right).
\end{equation}

\begin{table}
    \centering
    \resizebox{0.47\textwidth}{!}{
    \begin{tabular}{c c c c}
    \toprule
    & Methods & Hausdorff Distance  $\downarrow$ & Chamfer Distance  $\downarrow$  \\
	\midrule
		water mist & PCS-GAN	& 0.38	& 0.02  \\
		water mist & DERT \cite{huang2024sunshine}      & 5.81	& 1.56  \\
		smoke      & PCS-GAN	& 0.68	& 0.11  \\
	\bottomrule
    \end{tabular}}
    \caption{Realistic evaluation of generated random objects.}
    \label{tab:eval}
\end{table}

The objective function of $\mathcal{F}_p$ is as follows:
\begin{equation}
\begin{aligned}
	\mathbb{E}_{\mathbf{P}}\left[-\log D \left( P \right)\right]+\mathbb{E}_{\tilde{\mathbf{P}}}\left[-\log \left(1-D\left( \widetilde{P} \right)\right)\right]+ \\ \mathbb{E}_{\mathbf{P}}\left[-\log D_{\mathrm{S}}\left(P_{\mathrm{T}} \right)\right]+\mathbb{E}_{\tilde{\mathbf{P}}}\left[-\log \left(1-D_{\mathrm{S}}\left( \widetilde{P}_{\mathrm{T}} \right)\right)\right].
\end{aligned}
\end{equation}
During the training process, the sequence discriminator $D_S$ adopts the temporal discriminator loss~\cite{li2021tpu} of TPU-GAN. Discriminator $D$ uses improved WGAN loss~\cite{gulrajani2017improved}. In order to further reduce the local differences between the generated point cloud and the real point cloud, we use stitching loss~\cite{tang2022warpinggan} to enhance the generator $G$. When training PCS-GAN, we first fix the generator $G$ and PST-Net and update the discriminators $D$ and $D_S$. Then the discriminators $D$ and $D_S$ are fixed, and the generator $G$ and PST-Net are updated. Update each part alternately until the end of training.

\textbf{Realistic evaluation of generated random objects.} The similarity between generated and real data is evaluated using Hausdorff and Chamfer distances. Smaller distance values indicate higher similarity, as shown in Table\ref{tab:eval}. In our experiments, PCS-GAN was trained for 2000 epochs with a batch size of 4, and the initial sampling length $N$ of the point cloud sequence was set to 16.

\begin{figure}[!tb]
    \centering
    \includegraphics[width=1.0 \linewidth]{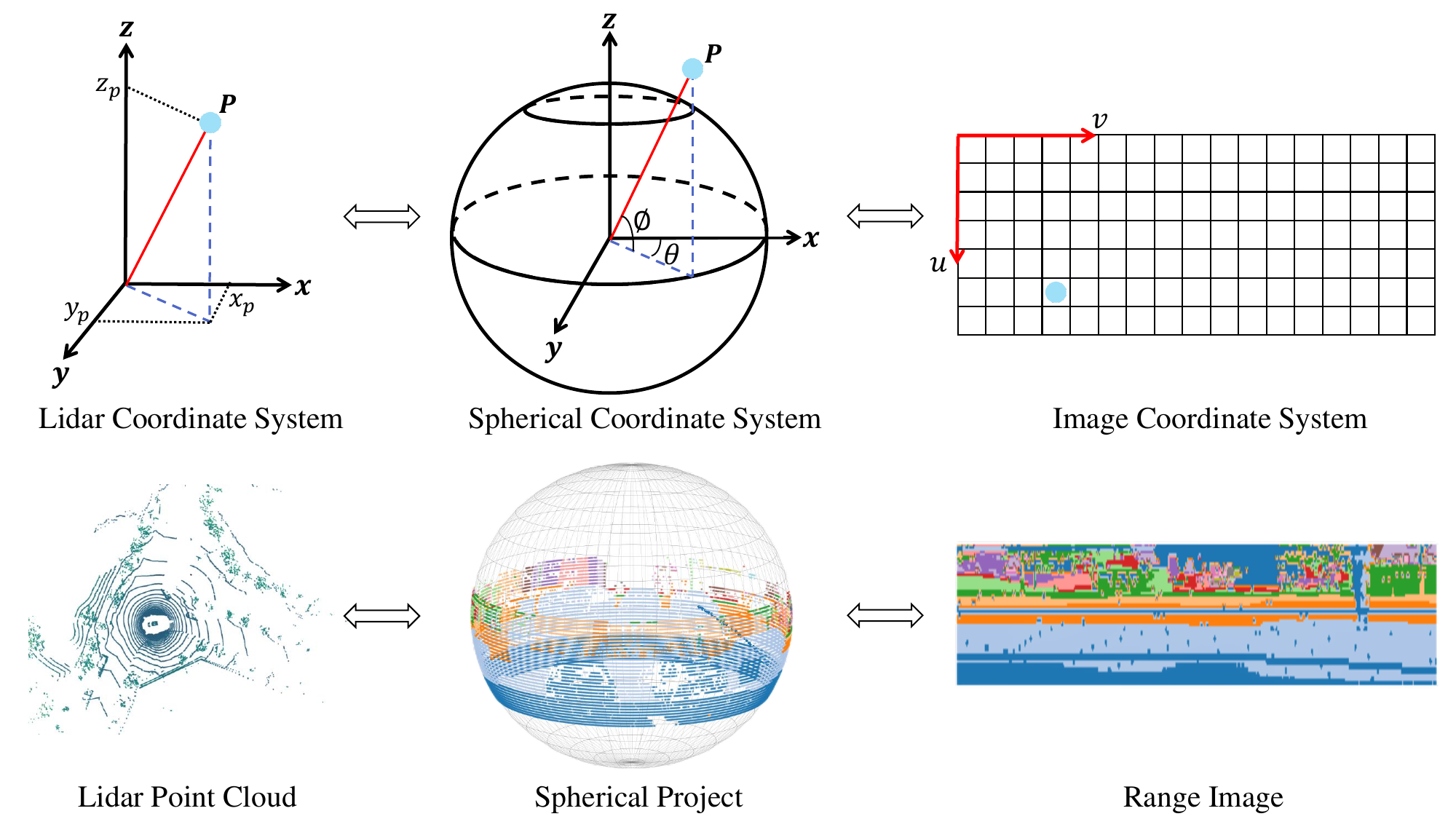}
	\caption{Obtaining a range image of the point cloud via spherical projection. The first line describes the three coordinate systems involved in the conversion process, while the second line outlines the changes in data format during this transformation.}
	\label{fig:LiDAR-sim}
\end{figure}

\subsection{Adversarial point cloud sequence generation}
We fuse the generated random object sequence and scene point cloud to form a new point cloud sequence. We use water mist as an example to describe the fusion process. First, align the coordinate system of the water mist with the coordinate system of the scene point cloud. Second, the water mist is attached to the body of the target vehicle under the coordinate system of the scene point cloud to complete the fusion. However, this destroys the inherent topological properties of LiDAR point clouds and also introduces occlusion issues. Therefore, we use a range image-based LiDAR simulation approach to decide which points are visible from the LiDAR perspective.

\textbf{LiDAR simulation.} The purpose of LiDAR simulation is to solve the occlusion problem caused by the fusion of random objects and point cloud objects in the scene, and to generate a fused point cloud with the real scanning characteristics of LiDAR. We implement LiDAR simulation using range image representation which provides physically accurate rendering as shown in Figure \ref{fig:LiDAR-sim}. Specifically, when we project the point cloud to the range image, if multiple points are projected to the same pixel at the same time, only the point closest to the LiDAR is retained. This simulates the occurrence of multiple points in the path of the laser beam, with later points being occluded by previous points. Subsequently, we back-project the range image to obtain the point cloud again. Therefore, the characteristics of a LiDAR scan are simulated during the projection process by replacing the depth values on the range image. We use this simulation approach to approximate the LiDAR renderer in this work. In addition, the range image has the problem of missing horizontal lines on KITTI. The detail of repair method can be found in supplementary material.

\textbf{Mix the water mist with car in point cloud.} After the coordinates are aligned, we select point $A$ of the water mist point cloud and move it to point $B$ of the car in the scene point cloud. Then, the water mist is adjusted according to the direction of the car to fit the body. The distance between the center of mass $O$ of the water mist point cloud and the point $A$ directly above is half the height $\Delta z$ of the point cloud. The height $\Delta z$ of the water mist point cloud is calculated as follows:
\begin{equation}
\Delta z=\frac{1}{2}{\left(\max\limits_{(x, y, z)\in P_{mist}} z - \min\limits_{(x, y, z)\in P_{mist}} z\right)},
\end{equation}
\noindent where $P_{mist}$ is the water mist point cloud. 

We first determine the top centers of the four bodies of the target vehicle, which are replaced by the midpoints of the four sides of the upper rectangle of the annotation box. Then take the top center of the car body facing the LiDAR as point $B$. However, vehicles do not always face the LiDAR head-on in real scenarios. Therefore, we set up four fusion modes, as follows:
\begin{itemize}
\item \textbf{head/tail side}: When the front or rear of the car faces the LiDAR, the water mist merges with the top center point $B$ of the front or rear of the car.
\item \textbf{body side}: When one side of the body faces the LiDAR, the water mist merges with the top center point $B$ of the body.
\end{itemize}

When multiple bodies (up to two) face the LiDAR:
\begin{itemize}    
\item \textbf{two sides}: Fuse water mist at the top midpoint of both bodies at the same time.
\item \textbf{corner point}: The top point $B$ of the intersection line of the two bodies merges with the water mist
\end{itemize}

\begin{table}
	\centering
	\resizebox{0.47\textwidth}{!}{
		\begin{tabular}{c c c c} 
				\toprule	
				\multirow{2}{*}{Dataset} & \multirow{2}{*}{Model} &  \multicolumn{2}{c}{Attack success rate}   \\
				\cmidrule(r){3-4}
				&  &  water mist & smoke\\	
				\midrule		
				\multirow{6}{*}{KITTI}  & TED ~\cite{wu2023transformation}             &  87.29   & 76.02   \\		
				& Focals Conv - F ~\cite{chen2022focal}        &  95.32   & 81.47   \\	
				& Part-A2-Anchor  ~\cite{shi2020points}        &  92.05   & 90.95   \\	
				& Part-A2-Free  ~\cite{shi2020points}          &  89.63   & 87.53   \\
				& PV-RCNN  ~\cite{shi2020pv}                   &  86.34   & 87.27   \\
				& PointPillar  ~\cite{lang2019pointpillars}    &  93.26   & 86.99   \\
				\midrule                        
				\multirow{6}{*}{nuScenes}  & VoxelNeXt   ~\cite{chen2023voxelnext}                   &  83.18   & 82.87   \\	
				& TransFusion-L  ~\cite{bai2022transfusion}               &  92.61   & 93.27   \\
				& CenterPoint(voxel\_size=0.1)  ~\cite{yin2021center}     &  86.67   & 87.60   \\
				& CenterPoint (voxel\_size=0.075) ~\cite{yin2021center}   &  85.06   & 85.73   \\
				& CenterPoint-PointPillar ~\cite{yin2021center}           &  86.26   & 81.43   \\
				& PointPillar-MultiHead                                   &  84.48   & 81.54   \\
				\bottomrule				
		\end{tabular}}
        \caption{Attack performance of 3D object detection models on different datasets under random object perturbations.}
	\label{tab:table1}
	\end{table}

\begin{table*}
	\centering
	\resizebox{\linewidth}{!}{
		\begin{tabular}{c c c c c c c c c c c}
			\toprule
			\multirow{2}{*}{Dataset}  & \multirow{2}{*}{Model}  & \multicolumn{9}{c}{Attack success rate}  \\
			\cmidrule(r){3-11}
			& &  left 40$^{\circ}$   & left 20$^{\circ}$   & left 10$^{\circ}$  & left 5$^{\circ}$  & 0$^{\circ}$   &  right 5$^{\circ}$  &  right 10$^{\circ}$   & right 20$^{\circ}$   & right 40$^{\circ}$  \\
			\midrule
			\multirow{6}{*}{KITTI} & TED ~\cite{wu2023transformation}         &25.02 &44.14 &54.12 &56.33 &55.74 &53.72 &53.44 &51.94 &31.79 \\
			& Focals Conv - F ~\cite{chen2022focal}    &28.42 &41.90	&57.39 &62.11 &62.03 &61.71	&58.33 &46.15 &29.01 \\
			& Part-A2-Anchor ~\cite{shi2020points}     &41.95 &53.05	&58.77 &60.64 &59.31 &57.86	&58.20 &56.28 &41.99 \\
			& Part-A2-Free ~\cite{shi2020points}       &33.50 &47.81	&53.83 &56.50 &55.02 &53.71	&52.07 &47.44 &32.43 \\
			& PV-RCNN ~\cite{shi2020pv}                &40.97 &46.60	&53.25 &55.44 &55.93 &56.61	&58.13 &55.18 &41.53 \\
			& PointPillar ~\cite{lang2019pointpillars} &56.62 &64.40	&70.72 &75.37 &74.69 &72.31	&70.04 &56.97 &49.94 \\   
			\midrule	     
			\multirow{6}{*}{nuScenes}  & VoxelNeXt   ~\cite{chen2023voxelnext}                   &76.34 &85.08 &90.84 &90.15 &91.44 &91.71 &91.90 &82.23 &75.14  \\	
			& TransFusion-L  ~\cite{bai2022transfusion}               &86.05 &88.59 &95.65 &95.83 &95.47 &96.20 &96.74 &94.02 &85.69  \\
			& CenterPoint(voxel\_size=0.1)  ~\cite{yin2021center}     &80.91 &88.91 &92.76 &93.23 &92.99 &92.53 &93.07 &89.30 &80.52  \\
			& CenterPoint (voxel\_size=0.075) ~\cite{yin2021center}   &82.43 &88.81 &91.17 &92.36 &91.73 &93.54 &93.93 &90.15 &84.79  \\
			& CenterPoint-PointPillar ~\cite{yin2021center}           &77.56 &82.69 &89.01 &92.31 &92.49 &92.67 &91.30 &87.00 &78.66  \\
			& PointPillar-MultiHead                                   &86.69 &88.34 &94.32 &94.94 &95.36 &96.28 &95.87 &92.88 &86.27  \\
			\bottomrule
	\end{tabular}}
        \caption{Effects of different angles of spraying water mist on attack performance.}
	\label{tab:table2}
\end{table*}

\textbf{Mix the water mist with car in range image.} We project the fused point cloud into range image. Because the points in the water mist may not appear on the laser path, we need to set an error limit in both the horizontal and vertical directions to determine whether the point is visible by LiDAR. Therefore, we use these two constraints, namely density parameters $d$, to control the density of random objects in the fused point cloud. In the experiment, the density parameter in the horizontal direction is denoted as $d_h$, and the density parameter in the vertical direction is denoted as $d_v$.

\section{Experiments}
\subsection{Experimental Setup}
\textbf{Dataset.} We evaluated the performance of adversarial attacks using two public autonomous driving datasets: KITTI~\cite{geiger2012we} and nuScenes~\cite{caesar2020nuscenes}. Since they do not provide labels for the test set, we use the training set and validation set for adversarial attack performance evaluation. For each dataset, we selected 3000 frame point clouds for the attack. Water mist perturbations were applied to each frame to create an adversarial point cloud sequence, and similarly, smoke perturbations were added to generate another adversarial sequence. In the experiments, the lengths of both the water mist and smoke sequences were set to 3.

\textbf{Attack Models.} We select high-performing 3D point cloud object detection models from each dataset for the attack evaluation. The models targeted for attacks on the KITTI include TED~\cite{wu2023transformation}, Focals Conv-F~\cite{chen2022focal}, Part-A2-Anchor~\cite{shi2020points}, Part-A2-Free~\cite{shi2020points}, PV-RCNN~\cite{shi2020pv}, PointPillar~\cite{lang2019pointpillars}. Among them, Focals Conv-F is a multimodal model that processes both images and point clouds as inputs. The models targeted for attacks on the nuScenes include VoxelNeXt~\cite{chen2023voxelnext}, TransFusion-L~\cite{bai2022transfusion}, CenterPoint(voxel\_size=0.1)~\cite{yin2021center}, CenterPoint (voxel\_size=0.075)~\cite{yin2021center}, CenterPoint-PointPillar~\cite{yin2021center}, PointPillar-MultiHead.

\textbf{Evaluation Metrics.} To evaluate the effectiveness of adversarial attacks involving random objects, we adopt the attack success rate as the primary metric. A successful attack is defined as a case where a vehicle initially detected by the object detection model is no longer detected after perturbations are applied. For detection to be considered successful, the vehicle confidence must exceed 0.5 and the intersection over union (IoU) between the predicted bounding box and the ground truth label must be greater than 0.7. In the experiment, the IoU threshold is set to 0.7.

\begin{figure}[!tb]
	\centering
	\includegraphics[width=1.0 \linewidth]{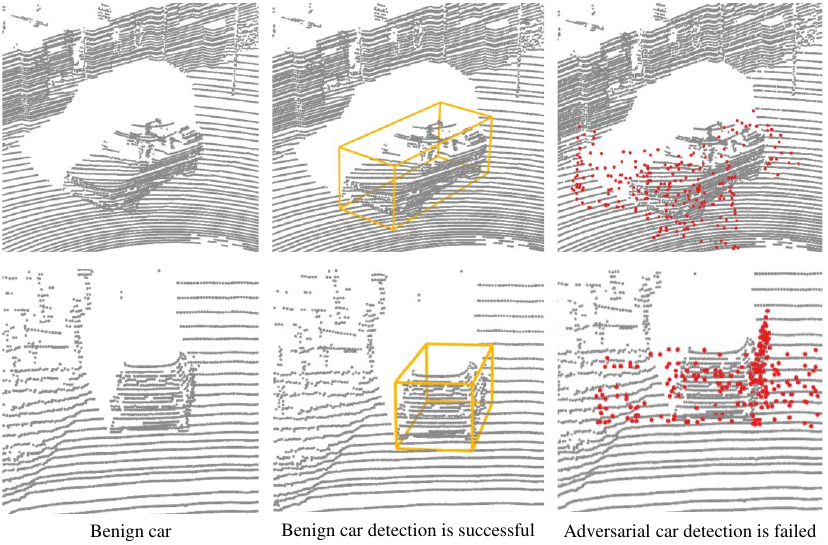}
	\caption{The performance of TED \cite{wu2023transformation} under the perturbation of water mist on KITTI when the fusion mode is two sides. Perturbed points of adversarial examples are marked in red.}
	\label{fig:result1}
\end{figure}

\begin{table*}
    \centering
    \resizebox{\linewidth}{!}{
        \begin{tabular}{c|ccc|ccc|ccc|ccc}
            \toprule
            \multirow{2}{*}{Model} 
            & \multicolumn{3}{c|}{Car AP@0.70, 0.70, 0.70} 
            & \multicolumn{3}{c|}{Car AP\_R40@0.70, 0.70, 0.70} 
            & \multicolumn{3}{c|}{Car AP@0.70, 0.50, 0.50} 
            & \multicolumn{3}{c}{Car AP\_R40@0.70, 0.50, 0.50} \\
            \cmidrule(r){2-4} \cmidrule(lr){5-7} \cmidrule(lr){8-10} \cmidrule(l){11-13}
            & easy & mod. & hard 
            & easy & mod. & hard 
            & easy & mod. & hard 
            & easy & mod. & hard \\
            \midrule
            PV-RCNN($\ast$) & 98.72 & 90.02 & 89.70 
                             & 99.31 & 94.78 & 94.42 
                             & 98.72 & 90.03 & 89.70 
                             & 99.31 & 94.78 & 94.42 \\
            PV-RCNN($\#$)   & 98.98 & 89.99 & 89.68 
                             & 99.40 & 94.88 & 94.15 
                             & 98.98 & 89.99 & 89.68 
                             & 99.40 & 94.88 & 94.15 \\
            \bottomrule
        \end{tabular}
    }
    \caption{Object detection performance with water mist-augmented point clouds.}
    \label{tab:robustness}
\end{table*}

\subsection{Results on KITTI and nuScenes Dataset}
We evaluated different 3D object detection models on the KITTI and nuScenes, with the results presented in Table \ref{tab:table1}. The value ranges of the density parameters of KITTI and nuScenes are $d_v\in[0, 0.004]$, $d_h\in[0, 0.5]$ and $d_v\in[0, 0.5]$, $d_h\in[0, 0.5]$ respectively. In the experiment, the fusion mode was set to two sides. For water mist perturbation, we set $d_v=0.004$, $d_h=0.5$ on KITTI, and set $d_v=0.08$, $d_h=0.08$ on nuScenes. For smoke perturbation, we set $d_v=0.002$, $d_h=0.25$ on KITTI, and set $d_v=0.05$, $d_h=0.05$ on nuScenes. We used 3000 frames of original point clouds and 3 frames of water mist or smoke sequences to generate 9000 mixed frames, which were then used for evaluating the models. From Table \ref{tab:table1}, the proposed method achieves high attack success rates on state-of-the-art models under different density parameters. For example, on the KITTI, the attack success rate ranges from a minimum of 86.34\% to a maximum of 95.32\% when the water mist density is high. Even when the smoke density is low, the attack success rate still reaches 93.27\% on the nuScenes. Moreover, the attack success rates of Focals Conv-F on water mist and smoke are 95.32\% and 81.47\%, respectively, demonstrating that the proposed attack method is also effective against multimodal models. In addition, Figure \ref{fig:result1} shows the attack performance of TED \cite{wu2023transformation} on KITTI under water mist perturbation when the fusion mode is two sides.

\subsection{Water mist spray angle simulation}
We simulated the impact of different water mist spraying angles on attack performance, as shown in Table \ref{tab:table2}. We took the center of mass of the water mist point cloud as the origin and rotated 0$^{\circ}$, 5$^{\circ}$, 10$^{\circ}$, 20$^{\circ}$, and 40$^{\circ}$ to the left and right respectively. The rotated 3 frame water mist sequence and 300 frames of the original point cloud were fused into an adversarial point cloud sequence and used for model evaluation. In the experiment, when the fusion mode is body side, we set $d_v=0.004$, $d_h=0.5$ on KITTI, and set $d_v=0.5$, $d_h=0.5$ on nuScenes. Experimental results show that the attack ability of water mist can be maintained when the rotation angle is within $10^{\circ}$, otherwise, the attack ability of water mist will decrease as the rotation angle increases.

\subsection{Effect of water mist augmentation}
We evaluated the effect of water mist augmentation on object detection performance, as shown in Table \ref{tab:robustness}. PV-RCNN($\ast$) refers to the model trained and tested on clean point clouds, while PV-RCNN($\#$) denotes training with both clean and water mist point clouds, with testing performed on clean point clouds. The results indicate a slight improvement in detection performance when the model is trained with water mist-augmented data, demonstrating that such augmentation effectively enhances the model's robustness and detection accuracy.

\subsection{Ablation study}
\textbf{Fusion Modes.} We verified the attack performance of different fusion modes, as shown in Table \ref{tab:table3}. We select 1000 frames of the original point cloud and 3 frames of water mist sequences for fusion on KITTI. Specifically, we set the density parameters $d_v=0.004$, and $d_h=0.5$, and the fused point cloud was evaluated on TED \cite{wu2023transformation}.  Obviously, the two sides mode has outstanding attack performance.

\begin{table}
    \centering
    \begin{tabular}{c c}
    \toprule
	Method & Attack success rate \\
	\midrule
		two sides        & 86.98 \\
		body side        & 61.01 \\
		head/tail side   & 55.30\\
		corner point     & 57.22 \\
	\bottomrule
    \end{tabular}
    \caption{Attack performance of TED \cite{wu2023transformation}under different fusion modes.}
    \label{tab:table3}
\end{table}

\textbf{Density parameters.}
We study the impact of density parameters on attack performance. When the fusion mode is two sides, we use 1000 frames of original point clouds and 3 frames of water mist sequences to generate fused point clouds on KITTI, which are used to evaluate TED \cite{wu2023transformation}. When fixing $d_h=0.5$ to verify the influence of $d_v$, and fixing $d_v=0.04$ to verify the influence of $d_h$, the results are shown in Table \ref{tab:table4}. The results show that as the density increases, the attack performance improves.

\begin{table}
    \centering
    \begin{tabular}{c c c}
		\toprule
		Direction & density & Attack success rate \\
		\midrule
		\multirow{5}{*}{$d_h$}  & 0.1  & 75.73 \\
		& 0.2  & 80.92 \\
		& 0.3  & 83.74 \\
		& 0.4  & 85.49 \\
		& 0.5  & 86.98 \\
		\midrule                    
		\multirow{3}{*}{$d_v$}  & 0.001  & 73.24 \\
		& 0.002  & 82.64 \\
		& 0.004  & 86.98 \\
		\bottomrule
    \end{tabular}
    \caption{Attack performance of TED\cite{wu2023transformation} under different density parameter values when the fusion mode is two sides.}
    \label{tab:table4}
\end{table}

\section{Conclusion} 
This paper introduces an adversarial attack framework that utilizes random objects to perturb targets in real scenarios, which can effectively attack state-of-the-art 3D object detection models. We used two types of advancing object perturbations, water mist and smoke, and verified their attack capabilities on LiDAR perception. We proposed a motion and content decomposition generative adversarial network, PCS-GAN, for point cloud sequence generation. To generate an adversarial point cloud sequence, we fuse the random object sequence produced by PCS-GAN with real point cloud data. In this process, we use a range image to simulate LiDAR scanning characteristics and effectively address the occlusion issue. The attack performance of the adversarial point cloud is verified on multiple autonomous driving datasets. Experimental results show that random objects have strong adversarial attack performance.

\section{Acknowledgments}
This work was supported in part by the National Natural Science Foundation of China (No. 62401225, 62471415), the Natural Science Foundation of Xiamen, China (No. 3502Z202472018), the Natural Science Foundation of Fujian Province, China (No. 2024J01115, 2023J01004), and the Jimei University Scientific Research Start-up Funding Project (No.ZQ2024034).

\bibliography{aaai25}

\clearpage
\appendix

\section{Appendix / supplemental material}

\subsection{Quantification of the effects of water mist on vehicles}
We investigated the influence of water mist disturbance on the automobile, predominantly manifested in alteration of point distribution pre and post obscuration by water mist. Using the front end of the car as a case study, Table \ref{suptab:table1} provides a quantitative description of the impact of water mist disturbance on the car. "Point1" represents the number of points on the original car, and "Point2" represents the number of points on the rear car disturbed by water mist.

Within a certain error range, at the same pressure, the closer the water mist is to the LiDAR, the lower the occlusion ratio, which shows that the penetration of water mist by the LiDAR is stronger. At the same distance, the greater the pressure, the higher the occlusion ratio, which means that the water mist obscures the car more seriously.

\begin{table}[h]
  \centering
  \resizebox{1.0\linewidth}{!}{
  \begin{tabular}{c c c c c}
    \toprule
    Water pressure & Distance & Points1  & Points2  & Occlusion ratio\\
    \midrule
    \multirow{4}{*}{0.45 Mpa}  & 2m    & 757   & 295  & 61.03\%    \\
                               & 4m    & 757   & 315  & 58.39\%    \\
                               & 6m    & 757   & 325  & 57.07\%    \\
                               & 8m    & 757   & 323  & 57.33\%    \\
    \midrule
    \multirow{4}{*}{0.50 Mpa}  & 2m    & 757   & 292  & 61.42\%    \\
                               & 4m    & 757   & 306  & 59.58\%    \\
                               & 6m    & 757   & 323  & 57.33\%    \\
                               & 8m    & 757   & 318  & 57.99\%    \\
    \midrule
    \multirow{4}{*}{0.55 Mpa}  & 2m    & 757   & 282  & 62.75\%    \\
                               & 4m    & 757   & 311  & 58.92\%    \\
                               & 6m    & 757   & 317  & 58.12\%    \\
                               & 8m    & 757   & 316  & 58.26\%    \\
    \midrule
    \multirow{4}{*}{0.60 Mpa}  & 2m    & 757   & 274  & 63.80\%    \\
                               & 4m    & 757   & 302  & 60.11\%    \\
                               & 6m    & 757   & 310  & 59.05\%    \\
                               & 8m    & 757   & 311  & 58.92\%    \\
    \bottomrule
  \end{tabular}}
  \caption{The quantitative description of the impact of water mist perturbation on the car.}
  \label{suptab:table1}
\end{table}

\begin{figure}[h]
	\centering 
	\includegraphics[width=1.0 \linewidth]{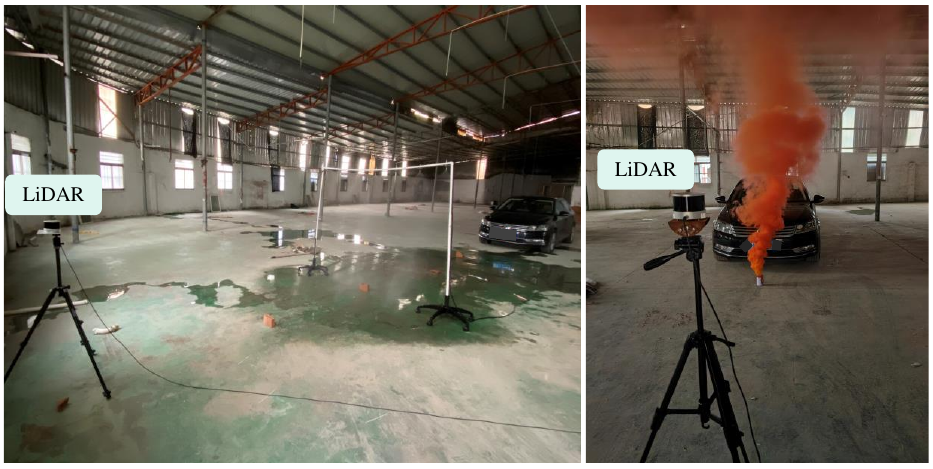}
	\caption{Data collection scenario of random objects. We collected LiDAR point cloud of sprayed water mist (left) and discharged smoke (right).}
	\label{fig:water}
\end{figure}

\subsection{Water mist spray angle simulation for different fusion mode}
To explore the relationship between attack performance and water mist spray angle, we also conducted experiments when the fusion mode is head/tail side. The experimental results are show in Table \ref{suptab:table2}

\begin{table*}[h]
	\centering
	\resizebox{\linewidth}{!}{
    \begin{tabular}{c c c c c c c c c c c}
		\toprule
		\multirow{2}{*}{Dataset}  & \multirow{2}{*}{Model}  & \multicolumn{9}{c}{Attack success rate}  \\
		\cmidrule(r){3-11}
		    & &  left 40$^{\circ}$   & left 20$^{\circ}$   & left 10$^{\circ}$  & left 5$^{\circ}$  & 0$^{\circ}$   &  right 5$^{\circ}$  &  right 10$^{\circ}$   & right 20$^{\circ}$   & right 40$^{\circ}$  \\
		\midrule
        \multirow{6}{*}{KITTI}  
                      & TED             & 31.95 & 53.52 & 55.78 & 55.15 & 55.62 & 54.87 & 54.47 & 52.26 & 29.10 \\
                      & Focals Conv - F & 30.07 & 62.50 & 63.99 & 64.74 & 65.17 & 64.86 & 62.85 & 59.08 & 32.27 \\
                      & Part-A2-Anchor  & 49.01 & 62.81 & 64.26 & 65.94 & 65.56 & 65.33 & 64.42 & 62.55 & 49.39 \\
                      & Part-A2-Free    & 39.98 & 55.80 & 58.84 & 59.74 & 59.61 & 59.37 & 58.67 & 55.27 & 38.58 \\
                      & PV-RCNN         & 38.55 & 58.24 & 60.09 & 59.98 & 59.98 & 60.20 & 59.22 & 57.41 & 40.33 \\
                      & PointPillar     & 50.77 & 72.27 & 73.30 & 73.02 & 72.82 & 72.67 & 71.55 & 71.16 & 50.22 \\

        \midrule
        \multirow{6}{*}{NuScenes}  
                      & VoxelNeXt                         & 55.25 & 74.40 & 79.74 & 81.03 & 82.50 & 82.23 & 81.68 & 79.10 & 60.31 \\
                      & TransFusion-L                     & 69.57 & 82.61 & 85.14 & 84.96 & 85.14 & 86.41 & 87.32 & 88.04 & 71.01 \\
                      & CenterPoint(voxel\_size=0.1)      & 62.59 & 84.76 & 85.99 & 86.22 & 86.99 & 87.61 & 87.99 & 87.84 & 63.43 \\
                      & CenterPoint (voxel\_size=0.075)   & 62.57 & 86.05 & 85.82 & 85.66 & 85.66 & 86.45 & 86.92 & 86.45 & 65.56 \\
                      & CenterPoint-PointPillar           & 71.34 & 88.00 & 90.57 & 91.30 & 90.75 & 90.84 & 90.75 & 91.21 & 73.08 \\
                      & PointPillar-MultiHead             & 76.99 & 84.83 & 87.41 & 87.62 & 86.89 & 86.17 & 86.07 & 86.89 & 76.06 \\

		\bottomrule
	\end{tabular}}
    \caption{Effects of different angles of spraying water mist on attack performance when the fusion mode is head/tail side.}
    \label{suptab:table2}
\end{table*}

\subsection{LiDAR Simulation}
Attaching random objects to target objects at the point cloud level can lead to ambiguity issues, such as multiple points being present along the same laser beam path. To address this, as described in the paper, we simulate LiDAR scanning using a range image. Specifically, when multiple points are projected onto the same pixel, only the point closest to the LiDAR sensor is retained. The range image is then back-projected to reconstruct a point cloud without ambiguity.

Ideally, LiDAR data can be seamlessly converted between point cloud and range image formats, as illustrated in Figure \ref{fig:LiDAR-sim} of the paper, without any data loss. However, during our experiments, we observed that the projection and back-projection process resulted in a loss of over $1\times 10^4$ points in the KITTI dataset. Given that each frame contains approximately $1.3\times 10^5$ points, this loss is significant. Further more, we noticed there were some horizontal missing in the obtained range image. To address these issues, we adopted the correction model proposed by \cite{bewley2020range}, which refines the LiDAR scanning model. Specifically, while maintaining the assumption that all laser beams are coplanar, the revised model no longer requires each beam to pass through the same origin. The m-line lasers be denoted as $l_1, l_2, ..., l_H$, then $l_i$ can be represented by the inclination angle $\theta_i$ and the height $h_i$ from the sensor origin. Therefore, the point $p=(x, y, z)$ be measured by $h_i$ and $d_{xy}$, as follows:
\begin{equation}
	d_{xy} = \sqrt{x^2 + y^2} \\
\end{equation}
\begin{equation}
	z = \tan \theta_i \cdot d_{xy} + h_i \\
\end{equation}

To estimate $\theta_i$ and $h_i$, we apply a scan unfolding technique to separate the data collected by different lasers. Linear fitting or Hough transformation is then used to estimate these parameters. Finally, all the parameters are incorporated into the repair model. After that, the range image back-project operation can be rewrite as follows:

\begin{equation}
	d'= \sqrt{d^2 - h_i^2\cos\theta_i} - h_i\sin\theta_i
\end{equation}
\begin{equation}
	x = d'\cos\theta_i \cos\phi
\end{equation}
\begin{equation}
	y = d'\cos\theta_i \sin\phi
\end{equation}
\begin{equation}
	z = d'\sin\theta_i + h_i
\end{equation}

\begin{table}
	\centering
	\begin{tabular}{c c c c c c}
		\toprule
		\diagbox{$d_v$}{$d_h$} & 0.1 & 0.2 & 0.3 & 0.4 & 0.5 \\
		\midrule
		0.001  & 33.94  & 41.76  & 46.03  & 47.35  & 49.74 \\
		0.002  & 44.14  & 50.23  & 52.04  & 54.20  & 55.10 \\
		0.003  & 47.73  & 52.08  & 53.93  & 55.59  & 54.91 \\
		0.004  & 49.28  & 52.53  & 53.85  & 55.37  & 55.93 \\
		\bottomrule
	\end{tabular}
        \caption{Attack success rate of PV-RCNN ~\cite{shi2020pv} under low-density water mist perturbation in body side mode.}
	\label{tab:table5}
\end{table}

\subsection{Ablation study}
\textbf{Lower limit of water mist density.}
Previous results have proven that fused point clouds improve attack capabilities as the density of random objects increases. Therefore, we also explored the attack performance when the water mist density is the lowest.  When the fusion mode is body side, we use 300 frames of original point clouds and 3 frames of water mist sequences to generate fused point clouds on KITTI, which are then used to evaluate PV-RCNN ~\cite{shi2020pv}. Under different water mist density parameters, the attack success rate of the fused point cloud on PV-RCNN ~\cite{shi2020pv} is shown in Tables \ref{tab:table5} and \ref{tab:table6}. Experiments show that low-density water mist has lower attack performance, which is also consistent with the impact of water mist density on targets in the real world.

\begin{table}
	\centering
	\begin{tabular}{c c c}
		\toprule
		$d_v$ & $d_h$ & Attack success rate \\
		\midrule
		\multirow{10}{*}{0.001}     & 0.01    & 13.38 \\
		& 0.02    & 18.33 \\
		& 0.03    & 21.16 \\
		& 0.04    & 24.49 \\
		& 0.05    & 26.04 \\
		& 0.06    & 27.85 \\
		& 0.07    & 29.86 \\
		& 0.08    & 31.59 \\
		& 0.09    & 33.67 \\
		& 0.10    & 33.94 \\
		\bottomrule
	\end{tabular}
        \caption{Attack success rate of PV-RCNN ~\cite{shi2020pv} under low-density water mist perturbation in body side mode.}
	\label{tab:table6}
\end{table}


\end{document}